\documentclass[10pt,letterpaper]{article}
\pdfoutput=1
\usepackage{cogsci}

\cogscifinalcopy % Uncomment this line for the final submission 

\usepackage{pslatex}
\usepackage{apacite}
\usepackage{float} % Roger Levy added this and changed figure/table
\usepackage[utf8]{inputenc} % allow utf-8 input
\usepackage[T1]{fontenc}    % use 8-bit T1 fonts
\usepackage{hyperref}       % hyperlinks
\usepackage{url}            % simple URL typesetting
\usepackage{booktabs}       % professional-quality tables
\usepackage{amsfonts}       % blackboard math symbols
\usepackage{nicefrac}       % compact symbols for 1/2, etc.
\usepackage{microtype}      % microtypography

\usepackage{graphicx}
\usepackage{amsmath}
\usepackage{amssymb}
\usepackage{wrapfig,lipsum}
\usepackage{enumitem}
\usepackage{subcaption}
\usepackage{latexsym,bm,amsmath, amssymb, amsthm, bbm, epsf, graphicx}
\usepackage{algorithm}
\usepackage[noend]{algpseudocode}

\usepackage{natbib}
\newcommand{\mainalgorithm}{%
\begin{algorithm}[H]
\caption{Switching semi-Markov jump model}\label{switchingJHMM}
\begin{algorithmic}[2]
\State \textit{\bf{Initialization}}
\State $O_{f}$ $\gets$ pre-processing
{\Comment{actions are observed with events}}
\State $O_{\rm tr}$, $O_{\rm te}$ $\gets$ cross-validation sequences for $O_{f}$
\State $B'$, $L'$ $\gets$ switchHMM($O_{\rm tr}$, $B$, $L$, criteria)
\State compute $A$ = GeneratorUpdate($B'$, $B$), $B = B'$, $L = L'$
\Repeat 
    \State \textit{\bf{Training}}
    \State $O$ $\gets$ TrajectorySampling($A$, $L$, $O_{\rm tr}$)
    \State $B'$, $L'$ $\gets$ switchHMM($O$, $B$, $L$, criteria)
    \State $B = B'$, $L = L'$
    \State \textit{\bf{Validation}}
    \State $ll_{\rm te}$ $\gets$ $P(O_{\rm te} | B, L)$
    \State recompute $A$ = GeneratorUpdate($B'$, $B$)
    \State validate structure of $A$
    \newline
    {\Comment{Make updates in the generator space}}
\Until{$ll_{\rm te}$ stops changing or max iterations reached}

\end{algorithmic}
\end{algorithm}
}
\title{Belief dynamics extraction}
\author{
  {\large \bf Arun Kumar}\\
  University of Minnesota\\
  Minneapolis, MN 55455 USA\\
  kumar250@umn.edu\\
  \And 
  {\large \bf Zhengwei Wu}\\
  Baylor College of Medicine\\
  Houston, TX 77030 USA\\ 
  zhengwei.wu@bcm.edu\\
  \AND 
  {\large \bf Xaq Pitkow}\\
  Rice University, Baylor College of Medicine\\
  Houston, TX 77030 USA \\ 
  xaq@rice.edu\\
  \And 
  {\large \bf Paul Schrater}\\
  University of Minnesota\\
  Minneapolis, MN 55455 USA\\
  schrater@umn.edu\\
  }
  
\begin{document}

\maketitle

\begin{abstract}
{
Animal behavior is not driven simply by its current observations, but is strongly influenced by internal states. Estimating the structure of these internal states is crucial for understanding the neural basis of behavior. In principle, internal states can be estimated by inverting behavior models, as in inverse model-based Reinforcement Learning. However, this requires careful parameterization and risks model-mismatch to the animal. Here we take a data-driven approach to infer latent states directly from observations of behavior, using a partially observable switching semi-Markov process.  This process has two elements critical for capturing animal behavior: it captures non-exponential distribution of times between observations, and transitions between latent states depend on the animal's actions, features that require more complex non-markovian models to represent. 
To demonstrate the utility of our approach, we apply it to the observations %actions
of a simulated optimal agent performing a foraging task, and find that latent dynamics extracted by the model has correspondences with the belief dynamics of the agent. 
Finally, we apply our model to identify latent states in the behaviors of monkey performing a foraging task, and find clusters of latent states that identify periods of time consistent with expectant waiting. This data-driven behavioral model will be valuable for inferring latent cognitive states, and thereby for measuring neural representations of those states.
}
\end{abstract}

%%%%%%%%%%%%%%%%%%%%%%%%%%%%%%%%%%%%%%%%%%%%
% STORYLINE
%%%
% Introduction
% - WHAT IS THE PROBLEM - why rate dependent MDP's for the foraging task? 
% - free-foraging paradigm
% - expectation? what's the expectation
% - Add an introductory figure explaining the problem statement
% - Behaviors are complex => complex belief space
% - Introduce the transfer to Data driven models
% 
% Background
% - different foraging models ??
% Model
% Experiment - toy, real, optimal
% Discussion 
% real data - belief representation/ what does it mean in terms of latent states
% optimal agent ? need to think
% Conclusion - WHAT WE CAN SAY ABOUT IT - need to think at the end?
%
%%%%%%%%%%%%%%%%%%%%%%%%%%%%%%%%%%%%%%%%%%%5
%\section{Quantifying latent states underlying behavior}
\section{Introduction}

% base animal behavior
% PROBLEM:
An animal's survival depends on effective planning for future costs and rewards. One of the most fundamental purposes of the brain is to create and execute such plans. %However, it is difficult to study these plans since they cannot be directly observed from behavior. 
However, these plans cannot be directly observed from behavior. To understand how the brain generates complex behaviors and learn how an animal builds a representation of the surrounding environment, it is valuable to construct hypotheses about the brain's internal states that narrow the search space for neural implementations of planning.  These hypotheses often come from models of the task implemented as artificial agents, whose internal state representations provided a latent space. However, differences between the model task and agent and the real task and animal create the potential for severe model-mismatch, injecting unknown biases into scientific conclusions. Here we use a latent-variable model to impute latent behavioral states based on observed behavior directly, using a data-driven latent-variable analysis that is designed to match the dependency structure of agent-based models without enforcing parametric structure. 

%% Need to fix this figure to show imputed states
\begin{figure}[ht]
\centering
  \includegraphics[width=1.0\linewidth]{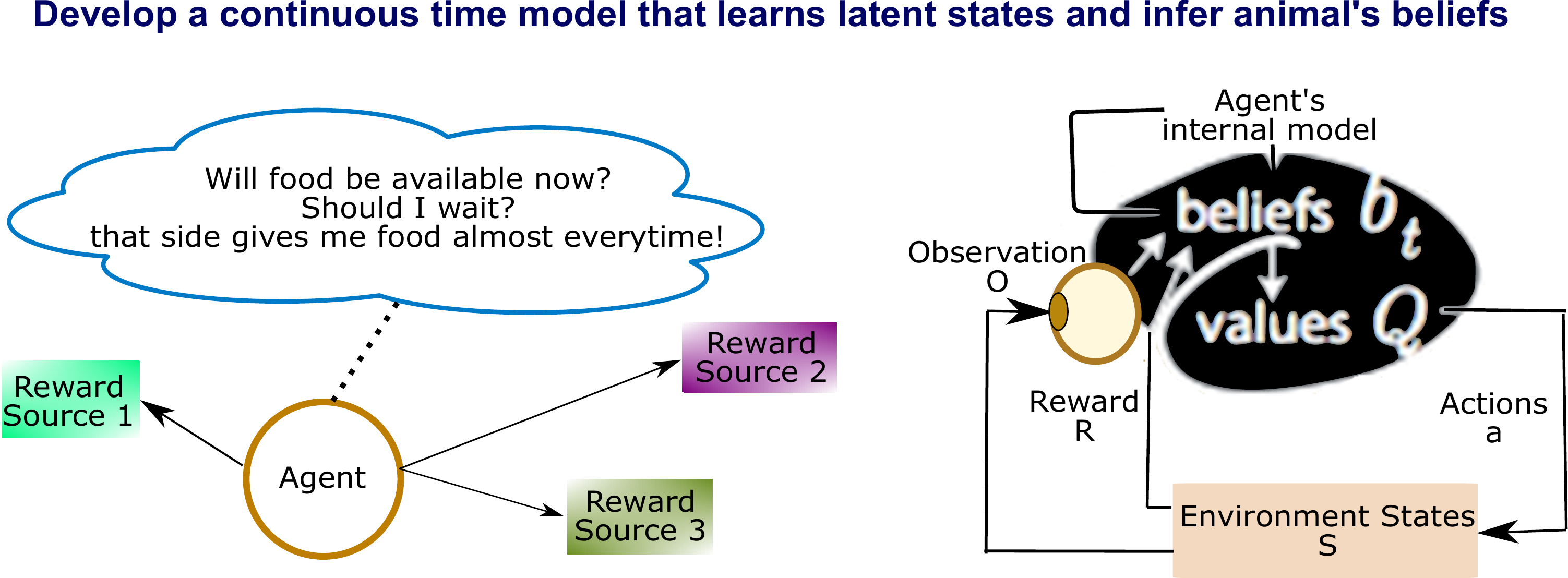}
\caption{{\it Overview}: In complex natural tasks such as foraging, an animal faces a continuous stream of choices. Some of the choices pertain to hidden variables in the world, such as food availability at a given location and time. These variables determine time- and context-dependent rates for observation events and rewards. To perform well at these tasks, animals must learn these hidden rates and act upon what they have learned. Our goal is to develop a data-driven, continuous-time model for inferring an animal's latent states and their dynamics.}
\label{fig:problem}
\end{figure}
To understand the mechanisms underlying behaviors, it is crucial to study hard tasks that involve inferring latent variables, since only then will an animal need to create a mental model of the world; otherwise the animal could perform well simply by responding to its immediate sensory input. Naturalistic foraging is one such task where an agent has to make decisions from many difficult choices in an uncertain environment. When foraging, an animal must take actions to procure rewards, and these actions have costs. How the animal schedules its actions determines the balance between total costs and rewards, \cite{charnov2006optimal}. The animal's goal in foraging is to use its energy resources for short term and long term sustenance. Decisions must be made continuously, and therefore time is a key ingredient in foraging: An animal benefits from tracking {\it when} reward is likely accessible at different locations. A natural way to represent such temporal quantities is in terms of dynamic event rates. For this reason, our work highlights the continuous-time aspects of decision problems.

Fig \ref{fig:problem} illustrates our motivation for the foraging problem. An agent develops an internal model and takes an action, which may result in a reward. As a result, the agent updates its internal model in an attempt to learn the environmental dynamics. We explore the plausibility that an animal's internal states in continuous time manifest as measurable consequences on its behavior, using a switching hidden semi-Markov model, and demonstrate the model's applicability in inferring latent states on a foraging task. 
% Foraging, Continuous time and free foraging

% What problem we are trying to solve, motivation

%In the remainder of the document, Section \ref{sec:background} provides background. In Section \ref{sec:Model}, we discuss the details of the presented model followed by the experiment and discussion in Sections \ref{sec:Experiment} and \ref{sec:Discussion}.

\section{Background}
\label{sec:background}

Behavior identification using computational models has a rich history, and clear value--the ability to learn rich representations of behavioral constituents provides important insights into underlying neural processes which can also be incorporated into the development of artificial agents (\cite{anderson2014toward}). Early behaviorists explored behavioral sequences in an attempt to learn determining causal factors underlying behavior, aiming to explain effects like when an agent switches to an alternate choice. These approaches are still common in animal ecology, where hidden Markov time series models (HMMs) have been used to analyse animal's internal states  \cite{nathan2008movement,langrock2012flexible}. \cite{macdonald1995hidden} proposed using  HMMs to capture causal structure in putative motivational states. However, they also observed that there are no one-to-one correspondences between the learned states and behavior, and \cite{zucchini2008modeling} found that behavior also influences internal states through feedback, challenging the dependency structure assumed by HMMs.  To capture non-stationarity in behavior, \cite{li2017incorporating} use temporally varying transition probabilities to model animal movement. However, behavior identification has struggled to produce more than a description of the behavior, with unknown relationships between the elicited latent states and the animal's representations. These failures are less surprising when it's realized the behavior expressible by HMMs is incompatible with key characteristics of observed behavior.   

In these works and others, an important question left unanswered is what kind of latent belief states could be inferred that not only represent belief dynamics but also the choices that an animal or an agent makes. We attempt to uncover latent state beliefs in a continuous time model and apply it to a complex ecological process, foraging, which has multiple underlying sub-processes including satisfaction of needs, searching for alternatives, motivation, decision making, and control.
We show that by generalizing allowing action-dependent transitions and more complex temporal dynamics, we can capture the expressivity of artificial agents designed for these domains, and highly interprable representations from animal behavior.

\section{Model}
\label{sec:Model}
%\subsection{Semi-Markov process}

Ecological behavior in animals is often well characterized by quick transitions between discrete behavioral modes. These transitions are difficult to predict from external events, and instead reflect a shift of the animal's internal state based on integrating events over a longer time scale. A process with quick transitions separated by long inter-event intervals can be approximated by a discrete-time hidden Markov process involving transition probabilities, but many of the probabilities (those for which the state is unchanged) will be close to one, while the remaining probabilities will be very small and decrease with the discrete time scale. Instead, we expect there will be advantages in treating these latent dynamics in {\it continuous time}, based on rates or time intervals between transitions and events.

A natural model to account for these point-like transitions in continuous time is the semi-Markov Jump Process, \cite{rao2013fast}. This process is a simple but powerful class of continuous-time dynamics featuring discrete states that transition according to a generator rate matrix, producing rich and flexible timing that is potentially better matched to animal behavior. In contrast, times of transitions between states in a Markov process are exponentially distributed, which describe animal behavior poorly.

%% Need to fix this figure to show imputed states
\begin{figure}[h]
\centering
  \includegraphics[width=1.0\linewidth]{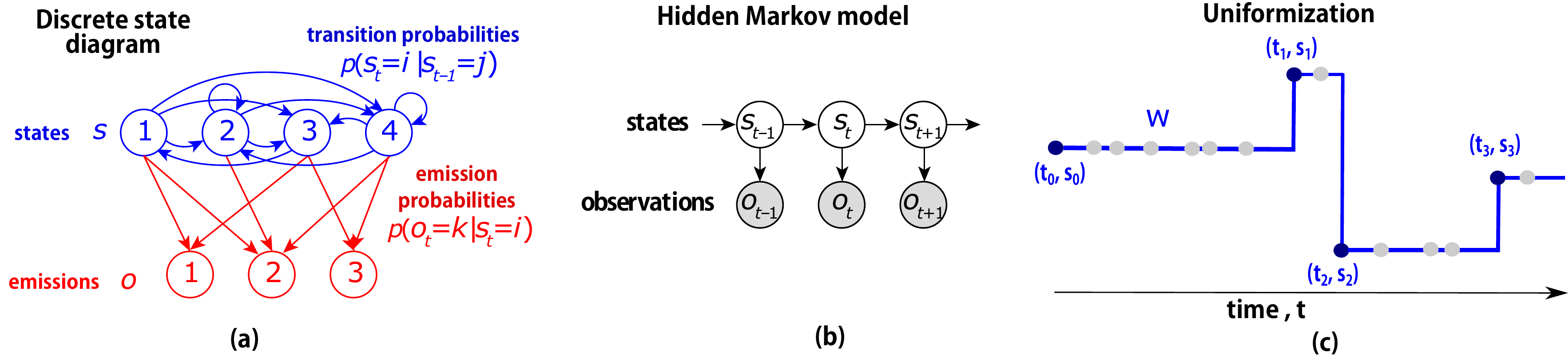}
\caption{A discrete-state Hidden Markov Model. {\it a}: Discrete state diagram shows latent states (blue circles) and their transitions (blue lines), as well as the possible emissions from each state (red circles) with their emission probability (red lines). {\it b}: Directed probabilistic graphical model showing dependence of state variable $s_{t+1}$ and observation $o_t$ on the previous state $s_{t}$. {\it c}: We present a continuous-time extension for latent states and discrete time observations using uniformization, \cite{rao2013fast}}
\end{figure}
%%

%To get around this severe limitation, we model latent states and observable states separately. 

However, agents who control their environment affect transition rates through their actions, which means a single generator rate matrix is not sufficient to model behavior. An important example are  Belief MDPs, which is a representation of a Partially Observable Markov Decision Process (POMDP, \cite{kaelbling1998planning}). POMDPs are a model for inference and control when sensory measurements provide only partial observations of the state of the world. Belief MDPs have distinct transition matrices that update beliefs differently for each action. Action-dependent transitions imply that a standard semi-Markov model with a single transition generator is not expressive enough to match action-dependent belief dynamics.

\begin{figure}[!h]
  \centering
  \includegraphics[width=0.95\linewidth]{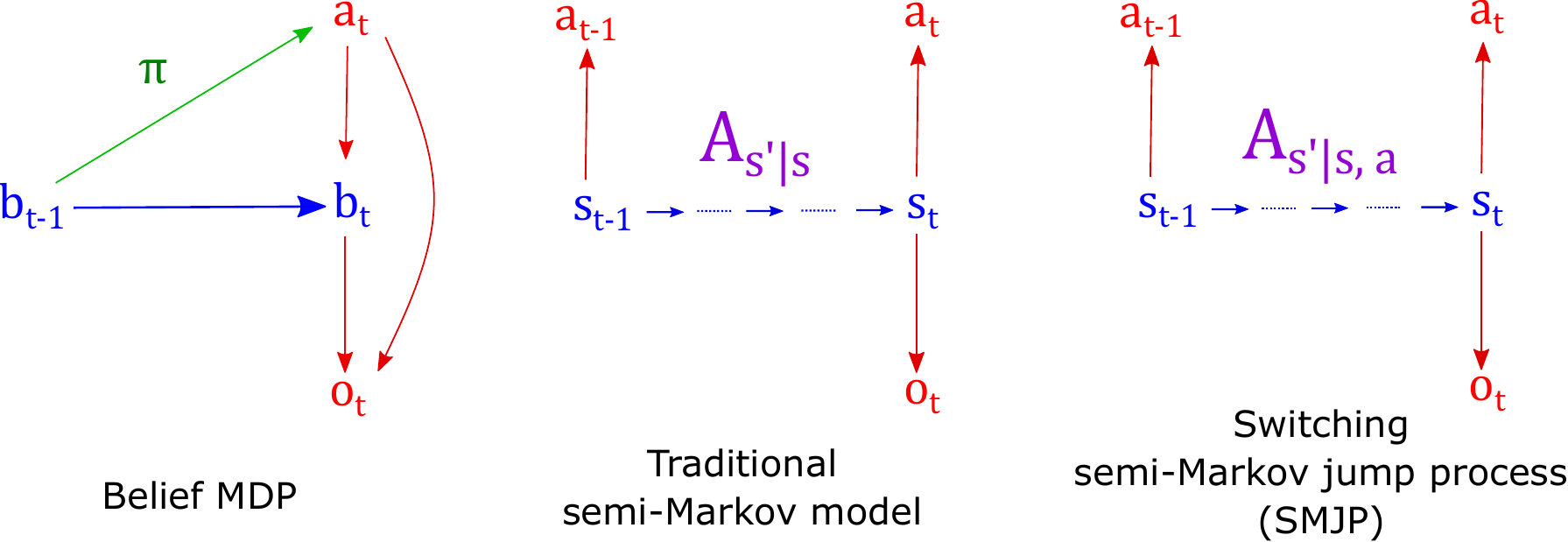}
  %\vspace*{-5mm}
  \caption{Comparison of graphical models of behavior. {\it Left}: In Belief MDP, belief transitions depend on actions selected by a policy. {\it Center}: Transitions in Semi-Markov Jump Process are independent of actions. {\it Right}: The Switching SMJP allows transition rates to depend on actions.}
 \vspace*{-3mm}\label{fig:switchSMJP}
\end{figure}

To allow for action-dependent belief dynamics, we propose a switching semi-Markov (SMJP) model that matches an agent's belief dynamics by switching its generator depending on the action $a$: $A_{s'|s,a}$. Let $s\in\mathcal{S}$ be a discrete latent state, and $A_{s'|s}$ be an ${N \times N}$ generator rate matrix that can be interpreted as an instantaneous transition matrix $A\,dt = P(s'(t+dt)|s(t))$.  This generator defines a point process that jumps from state $s$ to $s'$ at time $t$ according to the time-dependent matrix $P_t = \exp\left( A t \right)$.  The process can be implemented by sequentially sampling a time $t_i(s_i)$ from the total rate leaving state $s_i$, followed by sampling a new destination state $s'$ according to the matrix $P_{t_i(s_i)}(s'|s_i)$ evaluated at this sample time (Gillespie's algorithm). %To infer a behavioral model of a behaving agent based on experimental observations of its actions, we developed a continuous-time and switching variant of the forward-backward algorithm. We then inferred the transition and emission matrices through an Expectation-Conditional-Maximization algorithm. \cite{meng1993maximum}.
%A semi-Markov process for latent states, with observations emitted only occasionally and/or with some variability, is known as a Partially-Observed Semi-Markov Jump Process \cite{rao2013fast}. Although the latent states still exhibit exponentially-distributed interval distributions, the observations can appear with much more general interval distributions constructed as mixtures of exponentials. Let $s\in\mathcal{S}$ be a discrete latent state, and $A_{s'|s}$ be an ${N \times N}$ generator rate matrix that can be interpreted as an instantaneous transition matrix $A\,dt = P(s'(t+dt)|s(t))$.  This generator defines a point process that jumps from state $s$ to $s'$ at time $t$ according to the time-dependent matrix $P_t = \exp\left( A t \right)$.  The process can be implemented by sequentially sampling a time from an exponential distribution with rate equal to the total rate leaving state $s_i$, $t_i(s_i) \sim \exp\left(-\sum_{s'\neq s_i} A_{s'|s_i} \right)$, followed by sampling a new destination state $s'$ according to the matrix $P_{t_i(s_i)}(s'|s_i)$ evaluated at this sample time (Gillespie's algorithm \cite{gillespie1977exact}). 
An analogous process occurs for the generation of observable events $o$, through the emission generator matrix $B_{o|s}$. %In our case these observable emissions correspond to actions of the behaving agent.
%Within an interval $[t_{start}, t_{end}]$, the sampling probability is piece wise constant.
The resulting process is similar to a simple Markov process, except that the time between transitions is stochastic and depends on the starting state (but not the end state), illustrated in Fig \ref{fig:switchSMJP}; the animal's behaviors and decision making are continuous, albeit partially observable only at discrete recording times.
%%%%%%%
%\subsection{Fitting Semi-Markov Jump Processes}

%%%%% Start %%%%%%%%%%%%%%%%%%%

The Markov Jump Process %is one such method that 
extends discrete time Markov processes in continuous time. \cite{rao2013fast} introduced Markov chain sampling methods that simplify structures by introducing auxiliary variables. We adapt jump structures to provide a continuous-time representation for the free foraging task and the trajectory is introduced using a generator matrix. Let $A \in \mathbb{R}^{N \times N}$ be the generator matrix, which is skew symmetric and negative diagonal entries. We can represent 
$P_{t} \in \mathbb{R}^{N \times N}$  as continuous-time transition matrix given by $P_{t} = \exp(A t)$,  
$B_{t} \in \mathbb{R}^{N \times N}$ as discrete time transition matrix that is induced by {\em uniformization}, and $L_{t} \in \mathbb{R}^{N \times \left|O\right|}$ as observation matrix $P(O|s)$.

Uniformization instantiates the Markov Jump Process as a sequence of discrete time transition matrices, by introducing a latent sequence of random times that are adapted to the process generator but occur at a rate %slightly faster than the fastest transition $\Omega = (1+\epsilon) \max_s A_s$.
$\Omega \geq \max_s A_s$. For each interval, a random discretization vector of sampled times is $W = [w_1, w_2, ... , w_n]$, and we impute sampled times for a trajectory. Using this notation, we sample both random times as a Poisson process with intensity $\Omega$ and states using the generator matrix. The hidden Markov model characterizes a sample path of a piecewise constant stochastic process over these sampled and event times as $(s_0, S, T)$ where $T$ is now an ordered union of event times and randomly sampled discretized times. The chain can jump from a state to the same state or any other state, while the emissions are observed only at certain specified times.  Since we sample intervals with these virtual jump times, the constructed process represents the same chain.

To learn the discrete time transition matrix $B$ and emission matrix $L$, we consider an ensemble of sample sequence of observed emissions as generated from an HMM, and update the matrices using an EM algorithm to best account for the available observations. However, if we sample discrete times once, the estimates would be biased, so we resample latent trajectories repeatedly and randomly based on uniformization. The learned $B$ matrix is then used to update the generator matrix using the relation $A_{\rm new} = (B_{\rm new} - I) \Omega_{old}$ while preserving its structure, and the random times are resampled to adapt to the modified $A_{\rm new}$. The resulting algorithm exploits uniformization to enable learning the generator via an EM-algorithm, which is orders of magnitude more efficient than Gibbs sampling.

%
%Applying this learning approach allows us to identify distinct latent modes describing foraging behavior for both a simulated agent following a Partially Observable Markov Decision Process (POMDP), and for experimental data for monkeys performing a simple foraging task.

Belief MDPs are a convenient representation for POMDPs that treats current beliefs (posterior probabilities) over partially observable world states as fully observable. Agents following a Belief MDP exhibit transitions between beliefs $b_{t+1}=f(b_t,a_t,o_t)$, take actions according to a policy $\pi(a_t|b_t)$, and expect observations according to their beliefs via $p(o_t|b_t)$ (Fig \ref{fig:switchSMJP}). %A conventional SMJP is not expressive enough to capture the dynamics of a Belief MDP, because the SMJP has a {\it fixed} generator matrix $A_{s'|s}$, independent of action, whereas the Belief MDP has action-dependent belief dynamics. 
The switching SMJP model matches the agent's action-dependent belief dynamics by switching its generator conditional on the action $a$: $A_{s'|s,a}$. To infer the agent's model from experimental observations, we develop an EM algorithm to infer it's parameters. When applied to our switching model, the forward $\alpha$, backward $\beta$ and update $\xi$ equations of hidden Markov model, \cite{rabiner1989tutorial}, can be written as:
%
%\begin{align}
\begin{equation}
\begin{split}
\alpha_{t+1}^{k'}(j) = \left[\sum\limits_{i=1}^N \alpha_{t}^{k}(i) B_{ij}^{k}\right] L_j(o_{t+1}); \\
1 \leq t \leq T-1; 1\leq j \leq N; 1\leq k,k' \leq K
\end{split}
\end{equation}
%\end{equation}
%
\begin{equation} 
\begin{split}
\beta_t^k(i) = \sum\limits_{j=1}^N B_{ij}^k L_j(o_{t+1})\beta_{t+1}^{k'}(j); \\
t = T-1, T-2, ... ,1; 1 \leq i \leq N; 1\leq k,k' \leq K
\end{split}
\end{equation}
%\end{align}
%
where $k,k'$ are the action switching indices at time t and t+1 respectively. 
We adjust the model parameters to maximize the probability of the observation sequence given the model and train using EM. Updates are made using the $\xi$ variable, which is the probability of being in state $i$ at time $t$ and state $j$ at time $t+1$, and is given as
\begin{equation}
\begin{split}
\xi_t^k(i,j) = \frac{\alpha_t^k(i) B_{ij}^k L_j(o_{t+1}) \beta_{t+1}^{k'}(j)}{\sum\limits_{i=1}^N \sum\limits_{j=1}^N \alpha_t^k(i) B_{ij}^k L_j(o_{t+1}) \beta_{t+1}^{k'}(j)} ; \\
 1 \leq t \leq T-1; 1\leq i,j \leq N; 1\leq k,k' \leq K;
\end{split}
\end{equation}
%
%
%%%%%%%%%%%
\begin{figure}
    \centering
    \mainalgorithm
    \vspace*{-4mm}
    \caption{Overview of the algorithm.}
    \label{fig:algorithmOverview}
    \vspace*{-6mm}
\end{figure}
%\begin{figure}
    %\begin{minipage}[t]{0.65\textwidth}
    % need this comment symbol to avoid overfull hbox
    %\begin{subfigure}{1.0\textwidth}
    %    \mainalgorithm
    %\end{subfigure}
    %\end{minipage}
    % need this comment symbol to avoid overfull hbox
    %\begin{subfigure}{.45\textwidth}
    %    \includegraphics[height=5.5cm]{figures/sSMJPToyData.pdf}
    %    \caption{The log-likelihood of test data given a model trained with a given train size improves as we increase train size and stabilizes indicating stability of the model.}
    %    \label{fig:toydata}
    %\end{subfigure}
    %\begin{minipage}[t]{0.35\textwidth}
    %To-be updated.
    %\end{minipage}
%\end{figure}
%%%%%%%%%%%%

The usual semi-Markov model is a special case of the switching semi-Markov model where the generator remains the same without action dependent switching. Our model is a switching model that changes rate, transition and emission matrices in accordance with the action taken by the agent.

We learn the model using an EM approach %Expectation Maximization approach
, updating model parameters given transition times sampled by the uniformization process, and resampling the transitions given the new model parameters. Overview of the algorithm is shown in Fig \ref{fig:algorithmOverview}. The in-between times are sampled via trajectory in latent space, providing us a continuous time series that is then used for learning via Hidden Markov model. Upon learning the emission and transition matrices for a sampled set, we use scaling factor and make an update to the rate matrix while preserving its structure, re-sample to get a new continuous time sequence and learn emission, transition matrices again. This process is followed until the log-likelihood on held out data stops changing within a small tolerance.

%%%%%%%%%%%%%%%%%%%%%%%%%%%%%%%
\section{Experiment}
\label{sec:Experiment}
We perform three experiments. We use the simulated toy data both to estimate a required training size and to ensure that the switching model is able to learn latent states, establish correspondence between partially observable Markov decision process belief states with SMJP latent states using theoretical optimal agent model and, then, apply our method to a real agent in a free foraging task. The number of states were selected by estimating the value at which the log-likelihood on the validation set stops improving.
\subsubsection{Simulated toy data}
%
%\begin{wrapfigure}{r}{5.75cm}
%\vspace*{-3mm}
%\includegraphics[height=3.25cm]{figures/toyPlotLLFinal.pdf}
%\caption{The model is able to explain test data and log-likelihood on held out data starts flattening out at the true number of states.}
%\label{fig:toydata}
%\vspace*{-3mm}
%\end{wrapfigure} 
%
To create a toy test data generated by the assumed model, we set up two transition matrices and one emission matrix with 5 states, 2 emissions and observation 
dependent actions. The expected size of the output sequence is set to 5000. Initial 
action is selected randomly and based on the action index, a transition matrix is selected. Thus, the selected transition matrix and emission matrix combination is used to estimate state transition and generate an emission. The observations, times and actions are added to the output sequence and the observation dependent action value is updated to get new observations. The simulated toy data sequence is used as a basic check if the SMJP model can learn and explain the observations.
We fit switching SMJP model to the simulated data 
%(Fig \ref{fig:toydata}) and plot log-likelihood of the cross validation held-out data vs number of latent states. We 
and observe that the log-likelihood starts stabilizing as it reaches the true number of states. It means that the model is able to explain the test data with an equivalent number of latent states. Therefore, we pursue a similar procedure to estimate the required number of latent states for both the optimal agent and the real agent. 

%\begin{figure}[h]
%\begin{minipage}[t]{0.5\textwidth}
%\begin{subfigure}{0.9\textwidth}
%\centering
%\includegraphics[height=5cm]{figures/sSMJPToyData.pdf}
%  %\vspace*{-8mm}
%  \caption{The log-likelihood of test data given a model trained with a given train size improves as we increase train size and stabilizes indicating stability of the model.}
%  \label{fig:toydata}
%\end{subfigure}
%\end{minipage}
%%
%\begin{minipage}[t]{0.5\textwidth}
%    The log-likelihood of test data given a model trained with a given train size improves as we increase train size and stabilizes indicating stability of the model.
%\end{minipage}
%\end{figure}

\subsubsection{Optimal agent}
To test our SMJP model we fit it on an optimal agent performing a foraging task. We model the beliefs of an ideal observer in this task using a POMDP. %\cite{kaelbling1998planning}. %There is a one-to-one correspondence between a POMDP over partially observable world states $z$ and a fully observed Markov Decision Process over beliefs $b$ in which the `state' is the posterior distribution $b=p(s|o)$ over the world state $s$. 
There is a one-to-one correspondence between a POMDP over partially observable world states $z$ and a fully observed Belief MDP in which the `state' is the `belief' $b$ or posterior distribution $b_t=p(z_t|o_{1:t})$ over the world state $z$.%We use the latter formulation, defining belief states and the corresponding dynamics, to obtain the policy of the artificial agent. 
We solve this optimal actor problem using a Belief MDP on a discretized belief space. The agent keeps track of its belief state about the world following transition dynamics $p(b'|b, a)$, where $b'$ is the new belief state, $b$ is the current state, and $a$ is an action.  The agent's sensory information depends on the world state according to the probability $p(o|b,a)$. Upon taking action $a$, the agent receives immediate reward $R(b, b', a)$. The goal of the agent is to maximize the long-term expected reward $\mathbb{E}[\sum_{t = 0}^{t = \infty} \gamma^{\,t} R(b_t,b'_t,a_t)]$. Our model agent achieves this goal using a policy that solves for its policy  %\cite{bellman1957dynamic}, 
by value iteration on the discretized belief states.  

The beliefs serve as latent states which control the agent's behaviors, and give its actions a non-exponential interval distribution, which is recapitulated by the fitted switching SMJP. We find that the likelihood of the observed data is maximal for a number of states that is smaller than the true size of the underlying POMDP belief space, indicating that the semi-Markov process is able to compress the agent's dynamics into a smaller effective number of latent states.
To validate the semi-Markov model in our foraging task, we discover the latent states of the artificial agent for whom we know the ground truth. We model this agent as a near-optimal actor that maximizes reward given partial observations of the true process. This agent maintains beliefs about the availability of food at different locations. Our agent is suboptimal because we do not store the beliefs with arbitrary precision, but rather discretize the beliefs to a finite resolution, and allow some diffusion between those belief states.

\subsubsection{Application to the free-foraging task} 
We apply the SMJP model to infer latent states of agents performing a simple foraging task. We applied the model to both theoretical agents with near-optimal behavior, and real agents (macaques) whose behavior we measured experimentally. In this task, two boxes contained rewards that became available after random exponentially-distributed time intervals. If an agent presses a lever on one box when the food is available, that reward is released and that box timer is reset. The benefit of the reward is offset by two action costs: pressing the lever, and switching boxes. The state of the box is not observable, so the agent must choose its action based on an internal belief about the box, with the presumed goal of maximizing total reward minus costs. This internal belief constitutes a latent state that we infer using the semi-Markov process, both from the artificial agent and behaving monkeys.

We applied the SMJP model to infer latent states of macaques performing a simple two-box foraging task. The animal freely moved between two feeding boxes with levers that released food after an exponentially-distributed random time interval (mean of 10 or 30 sec) had passed.  %(mean of 15 or 25 sec)
The model observations were lever pressing, reward delivery, and location within the box (Fig \ref{fig:monkeybehavior}a). Actions were: stay, move, or press either lever. The monkey's movements were tracked using overhead video, and quantized by $k$-means into different locations. The number of latent states %, 11, for the optimal agent as well as the real agent, 
is estimated by the log-likelihood maximization (Fig \ref{fig:monkeybehavior}b). The resultant process constructs the monkey's latent states to explain the non-exponentially-distributed intervals between lever presses (Fig \ref{fig:monkeybehavior}).
%
%%%%%%%%%%%%%%%%%%%%
% Interpretation of latent States
\section{Results and Discussion}
\label{sec:Discussion}
\subsubsection{Optimal agent}
%\textbf{Optimal Agent}
We trained the SMJP on an observation sequence generated by the optimal agent, and optimized the number of SMJP latent states by maximizing the log-likelihood of held-out data. While the Belief MDP agent's relevant states $Z$ (including location, reward, and beliefs $b$) should be implicitly embedded in the SMJP latent states $s$, these two state representations are not immediately comparable.

To establish a correspondence, we compute the joint distribution over $s$ and $Z$ at any one time point using the shared time series of observations: $p(s,Z|obs)=\tfrac{1}{T}\sum_t p(s_t|o_{1:T})p(Z_t|o_{1:T})$. This joint distribution shows which SMJP and POMDP states tend to occur at the same time. %For better interpretation of possible correspondences, we would like to know what $s$ states mean in terms of $Z$ states and gain some insights into encoding of $Z$ states into $s$ states.
It therefore provides a dictionary for translating the interpretable POMDP $Z$ states into our learned and unlabeled SMJP $s$ states.
\begin{figure}[!h]
\includegraphics[width=\linewidth]{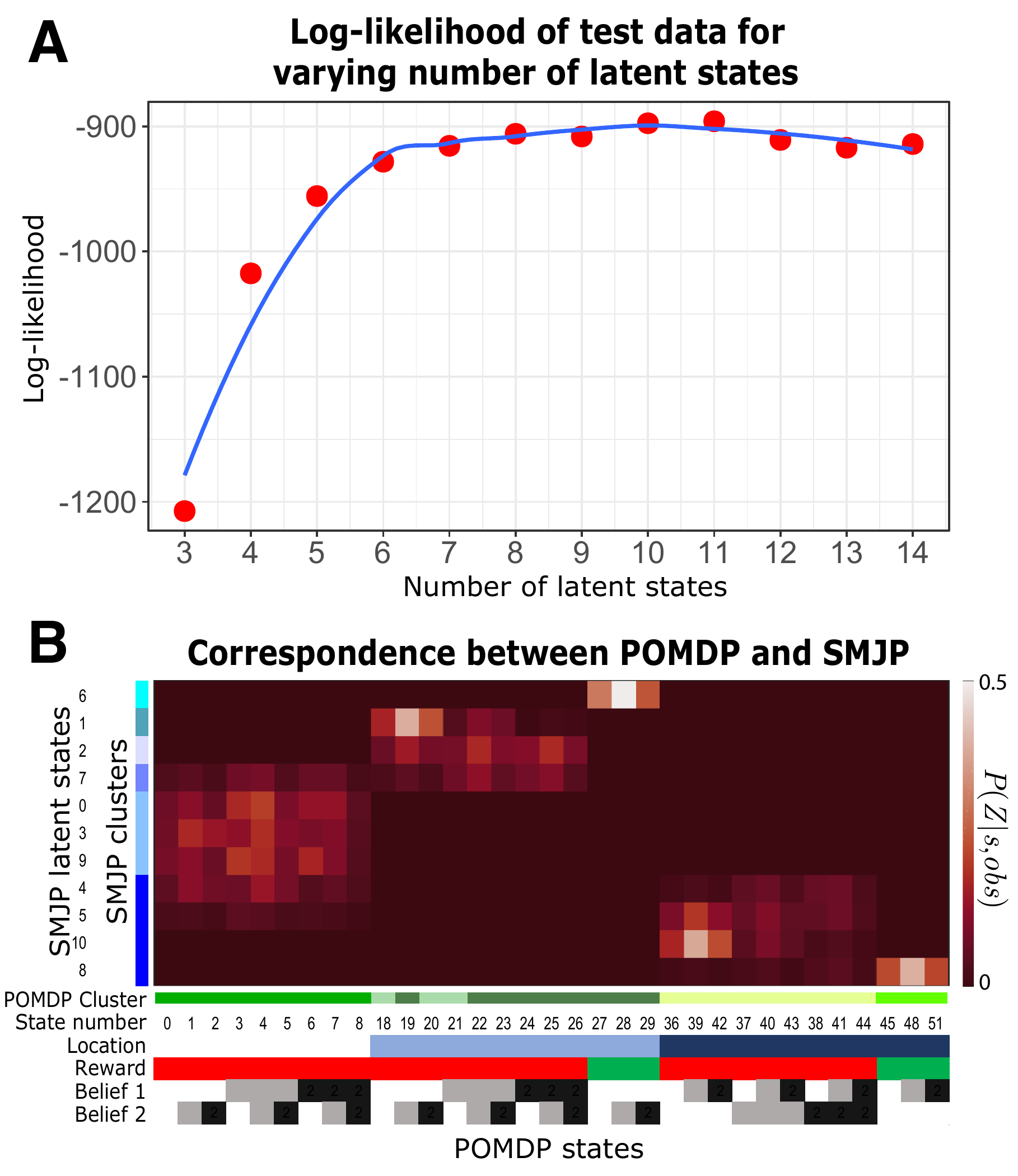}
    %\vspace*{-2mm}
    %\begin{subfigure}{1.0\linewidth}
    %    \centering
    %    \includegraphics[width=0.7\linewidth,height=%3.55cm]{figures/optLLPlotFinal.pdf}
    %    %\vspace*{-2mm}
    %    \caption{}
    %\end{subfigure}
    %\begin{subfigure}{1.0\linewidth}
    %    \centering
    %    \includegraphics[width=0.95\linewidth,height%=4.75cm]{figures/optBeliefSMJPCorrespondences11statesMixedBeliefs.pdf}
    %    %\vspace*{-2mm}
    %    \caption{}
    %\end{subfigure}
    %\\
  \caption{
   {\it{Latent states inferred by SMJP for an optimal agent implementing a POMDP}}. ({\bf a}) Log-likelihood on held out data provides an estimate of the required number of latent states. %{\it b)} Information contained in SMJP and POMDP co-clusters of P(Z|s,obs). {\it c)} %Correspondence of POMDP belief states with SMJP latent states: The SMJP latent states associate with POMDP hybrid belief states revealing the existence of the latent states dynamics that match agent's belief dynamics. We identify the number of required co-clusters by finding a combination that gives minimum error or loss in the computed mutual information. Existence of correspondences between latent states in Belief MDP and SMJP means that SMJP latent states dynamics structure matches optimal agent's belief dynamics structure.
   ({\bf b}) Co-clustering of states in a POMDP and our SMJP, based on the conditional probability of observing each POMDP state $Z$ from each SMJP state, $P(Z|s,obs)$. The POMDP states $Z$ are depicted below the horizontal axis. Clustered structure in the plot reveals that the SMJP states have information about the agent's belief dynamics.
   }
 \vspace*{-3mm}\label{fig:optBeliefStates}
\end{figure}
%
%% 
%Since the SMJP latent states implicitly learn the world model, we examine their correspondence with the hybrid POMDP belief states. While there may not be any obvious one-to-one mapping between the SMJP latent states, $s_t$, and the POMDP belief states, $z_t$, nonetheless both states are influenced by the shared time series of observations, according to $p(s,z)=\sum_t p(s_t|o_{1:T})p(z_t|o_{1:T})$. For the test observation sequence, the SMJP states are computed using trained model matrices and the forward-backward algorithm. Hybrid POMDP states are composite states of location , reward and two belief states. To assess correspondences with beliefs, we condition on observable i.e. location and reward to get $p(s, z | obs)$. The SMJP latent states do not have any labels attached to them but the POMDP states have clear interpretations. 

To increase interpretability, we cluster $p(Z|s, o)$ using information theoretic co-clustering, \cite{dhillon2003information}, which provides a principled coarse-graining of the states with improved semantic interpretability. We determine the required numbers of SMJP and POMDP co-clusters by finding minimum information loss in information theoretic co-clustering. Fig \ref{fig:optBeliefStates}b shows that latent SMJP states are associated with different belief states. Co-clustering also reveals that the SMJP latent states have dynamics that match the belief dynamics (not shown). These results demonstrate that the switching SMJP model can capture latent belief states and dynamics for behavioral data.

\subsubsection{Real agent}
%\textbf{Real agent}
%

\begin{figure*}[h]
  \vspace*{-5mm}
    \centering
    \includegraphics[width=0.95\linewidth,height=9.65cm]{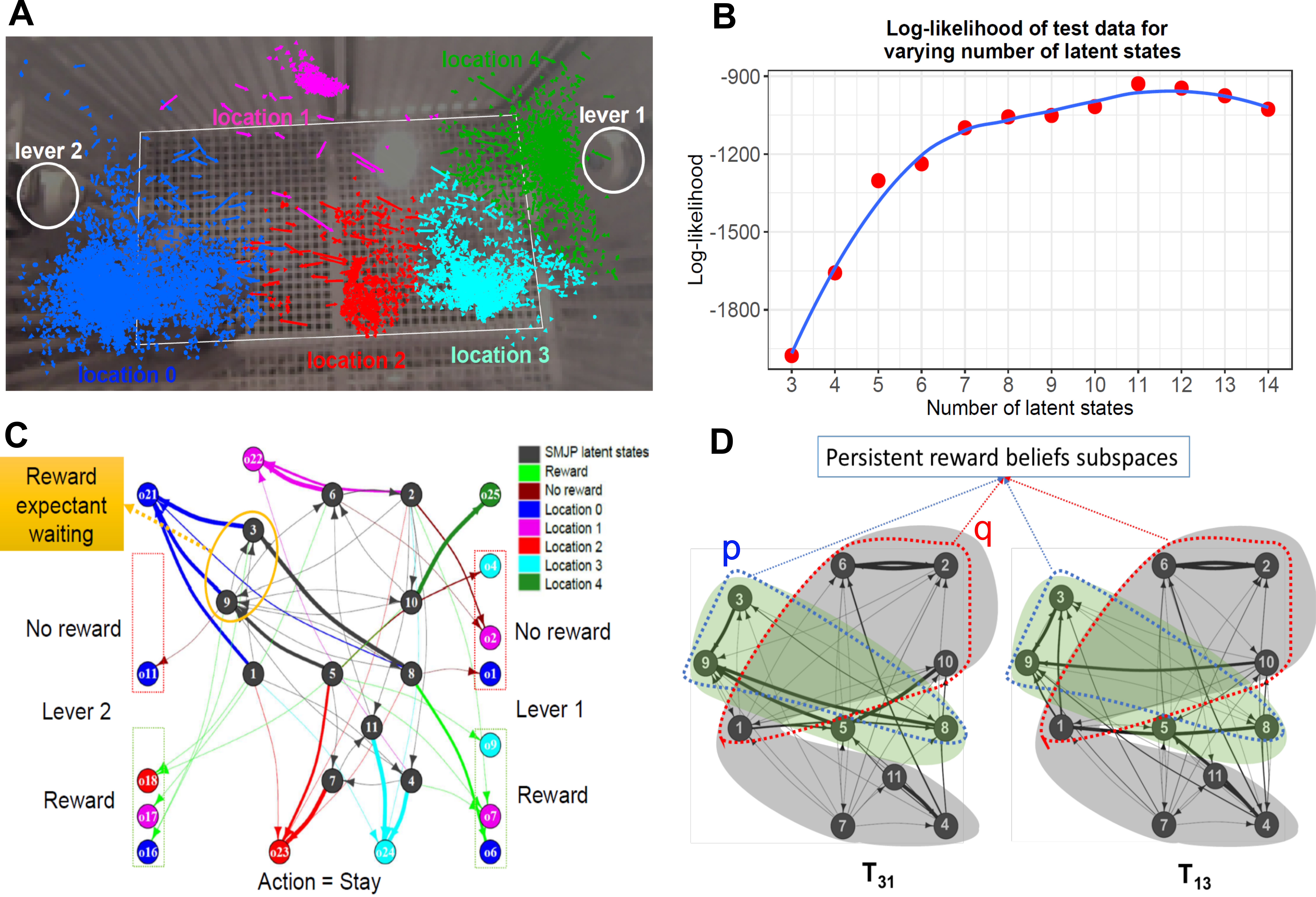}
  \vspace*{-2mm}
  \caption{Analyzing behavioral data from a freely moving monkey using the SMJP. ({\bf a}) Overhead video (background image) tracked the locations and normalized velocities (vectors) of the monkey. These data were then clustered by the $k$-means algorithm. ({\bf b}) We get an estimate of the required number of latent states by observing log-likelihood on held out data. ({\bf c}) SMJP model for observed monkey behavioral data for the action stay. %The latent states are numbered 1-11, labels: o1-o25 represent observations and lever press observations with or without reward are shown as dotted rectangular boxes. %All experiments were performed under protocols approved by The University of Texas at Houston Animal Care and Use Committee.
  Highlighted reward expectant waiting states illustrate that the latent states as regressors for the beliefs dynamics are useful in understanding monkey's behavior.
  ({\bf d}) Subspaces $p$ and $q$ (blue and red dotted), within the subgraphs (green and gray highlighted) for the joint operators $T_{31}$ and $T_{13}$ reveal persistent reward belief states. 
 }
 \vspace*{-5mm}\label{fig:monkeybehavior}
\end{figure*}
%
%We applied the SMJP model to infer latent states of macaques performing a simple two-box foraging task. The animal freely moved between two feeding boxes with levers that released food after an exponentially-distributed random time interval (mean of 15 or 25 sec) had passed. The SMJP model observations were lever pressing, reward delivery, and location/movements within the box. Actions were: stay at a location, move, or press either lever. The monkey's movements were tracked using overhead video, and quantized by $k$-means. 

The SMJP model constructs latent states and dynamics using the real agent's observations to predict choices and timing, including the non-exponentially-distributed intervals between lever presses. Fig \ref{fig:monkeybehavior}c shows states extracted for the action 'stay'. Beliefs precede an action and the extracted states reflect beliefs for the next action. For example, being in states $5,8$ are rewarding to the monkey. States that can be interpreted as `expectant waiting for reward' are highlighted (Fig \ref{fig:monkeybehavior}c): these states form a self-exciting delay network that is activated from other rewarded belief states. Moreover, the lower entropy of latent states associated with box 1 %(1: 0.3878, 2: 0.6428) 
revealed guarding behavior we identified from video. Overall, the model network encodes a set of complex but interpretable dynamics of the animal's beliefs and reward expectations which emphasize the complex computations underlying the decision making process. %The extracted latent states and dynamics will be useful regressors for finding neural correlates of the computations underlying the monkey's behavioral dynamics.

Each transition matrix acts like an action operator and the real agent performs operations in sequences. So, we examine joint operators $T_{ji} = T_i  T_j$, where $T_i$ and $T_j$ are operators for actions $i$ and $j$ respectively. We use an off-the-shelf package using, \cite{brandes2008modularity} to extract subgraphs and then persistent subspaces from all the six joint operators corresponding to different action pairs. Fig \ref{fig:monkeybehavior}d shows subgraphs for two joint operators of interest (involving actions: lever press and stay). %: lever press followed by stay ($T_{13}$) and stay followed by lever press ($T_{31}$). 
The latent states (within subspaces $p$ and $q$) appearing in the same subgraphs of the joint operators illustrate the real agent's persistent reward belief states. %The latent states in subspace $p$ appear within the same subgraphs in all but $T_{12}$ joint operator. This joint operator implies the agent's intentions of taking the action 'move' following the action 'stay'. This shift in states in subspaces relates to the intention of moving than a belief of reward. %, exemplifying that the extacted states could capture subtle belief dynamics. %The states within the blue dotted lines appear within the same subgraphs in all but one ($T_{1,2}$) operator. It is explainable because operator $T_{1,2}$ means the agent is going from moving to stay mode and does not seem to involve intentions for lever press. 
The states outside the subspaces $p$ and $q$ correspond to other beliefs, for example, switching. These results demonstrate that the presented model is able to extract subtleties, albeit complex, in the belief states and their dynamics. The extracted latent states and dynamics will be useful regressors for finding neural correlates of the computations underlying the monkey's behavioral dynamics.

\section{Conclusion}

We presented a continuous-time switching semi-Markov model that learns the latent states dynamics in conformance with the belief structure of a partially observable Markov decision process. The revealed latent states are capable of inferring complex animal behavior and its belief dynamics in naturalistic tasks like foraging. Several aspects of the inferred behaviors and belief dynamics were examined to reveal that indeed, the internal latent structural representation match the agent's belief structure. The data-driven switching semi-Markov model provides useful estimates of the structure of the internal latent states for hard tasks. The latent states from this behavioral model could potentially be used to understand correspondences between neural activity and the latent belief dynamics that govern how an animal selects actions.

\subsubsection*{Acknowledgments}
The authors thank Dora Angelaki, Valentin Dragoi, Neda Sahidi and Russell Milton for useful discussions. AK, ZW, XP and PS were supported by BRAIN Initiative grant NIH 5U01NS094368.

\bibliographystyle{apacite}

\setlength{\bibleftmargin}{.125in}
\setlength{\bibindent}{-\bibleftmargin}

\vspace*{-2mm}
\bibliography{References}

\end{document}